\documentclass[10pt,twocolumn,letterpaper]{article}

\usepackage{iccv}
\usepackage{times}
\usepackage{epsfig}
\usepackage{graphicx}
\usepackage{amsmath}
\usepackage{amssymb}
\usepackage{gensymb}
\usepackage{enumitem}

\usepackage{multirow}
\usepackage{booktabs}
\usepackage{graphicx} 
\usepackage{tabularx} 
\usepackage{array,multirow,graphicx}
\usepackage{tabu}

\usepackage{caption}
\usepackage{subcaption}
\usepackage{pdfpages}

\usepackage{xfrac}

\newcommand{\tb}[1]{\textbf{#1}}
\newcommand{\mb}[1]{\mathbf{#1}}
\newcommand{\subsec}[1]{\vspace{3pt}{\setlength{\parindent}{0pt}\textbf{#1}}\hspace{3pt}}
\graphicspath{{./img/}}  

\usepackage[pagebackref=true,breaklinks=true,letterpaper=true,colorlinks,bookmarks=false]{hyperref}

\iccvfinalcopy 

\def\iccvPaperID{4237} 

\ificcvfinal\fi
\begin{document}
\title{
OmniMVS: End-to-End Learning for Omnidirectional Stereo Matching
%
}
%
\author{Changhee Won, Jongbin Ryu and Jongwoo Lim\thanks{Corresponding author.} \\
\small Department of Computer Science, Hanyang University, Seoul, Korea. \\
{\tt\small \{chwon, jongbinryu, jlim\}@hanyang.ac.kr}
}

\maketitle
\thispagestyle{empty}

\begin{abstract}

In this paper, we propose a novel end-to-end deep neural network model for omnidirectional depth estimation from a wide-baseline multi-view stereo setup.
The images captured with ultra wide field-of-view (FOV) cameras on an omnidirectional rig are processed by the feature extraction module, and then the deep feature maps are warped onto the concentric spheres swept through all candidate depths using the calibrated camera parameters.
The 3D encoder-decoder block takes the aligned feature volume to produce the omnidirectional depth estimate with regularization on uncertain regions utilizing the global context information.
In addition, we present large-scale synthetic datasets for training and testing omnidirectional multi-view stereo algorithms.
Our datasets consist of 11K ground-truth depth maps and 45K fisheye images in four orthogonal directions with various objects and environments.
Experimental results show that the proposed method generates excellent results in both synthetic and real-world environments, and it outperforms the prior art and the omnidirectional versions of the state-of-the-art conventional stereo algorithms.
\vspace{-10pt}
\end{abstract}

\section{Introduction}
%
%
Image-based depth estimation, including stereo and multi-view dense reconstruction, has been widely studied in the computer vision community for decades.
In conventional two-view stereo matching, deep learning methods~\cite{ilg2018occlusions,chang2018pyramid} have achieved drastic performance improvement recently. 
%
Besides, there are strong needs on omnidirectional or wide FOV depth sensing in autonomous driving and robot navigation to sense the obstacles and surrounding structures.
Human drivers watch all directions, not just the front, and holonomic robots need to sense all directions to move freely.
However, conventional stereo rigs and algorithms cannot capture or estimate ultra wide FOV ($>\!\!\!180^\circ$) depth maps.
Merging depth maps from multiple conventional stereo pairs can be one possibility, but the useful global context information cannot be propagated between the pairs and there might be a discontinuity at the seam.

%
%
%

\begin{figure}[t!]
  \begin{subfigure}[b]{0.356\linewidth}
  \centering
  \captionsetup{justification=centering}
  \includegraphics[keepaspectratio=true,width=\linewidth]{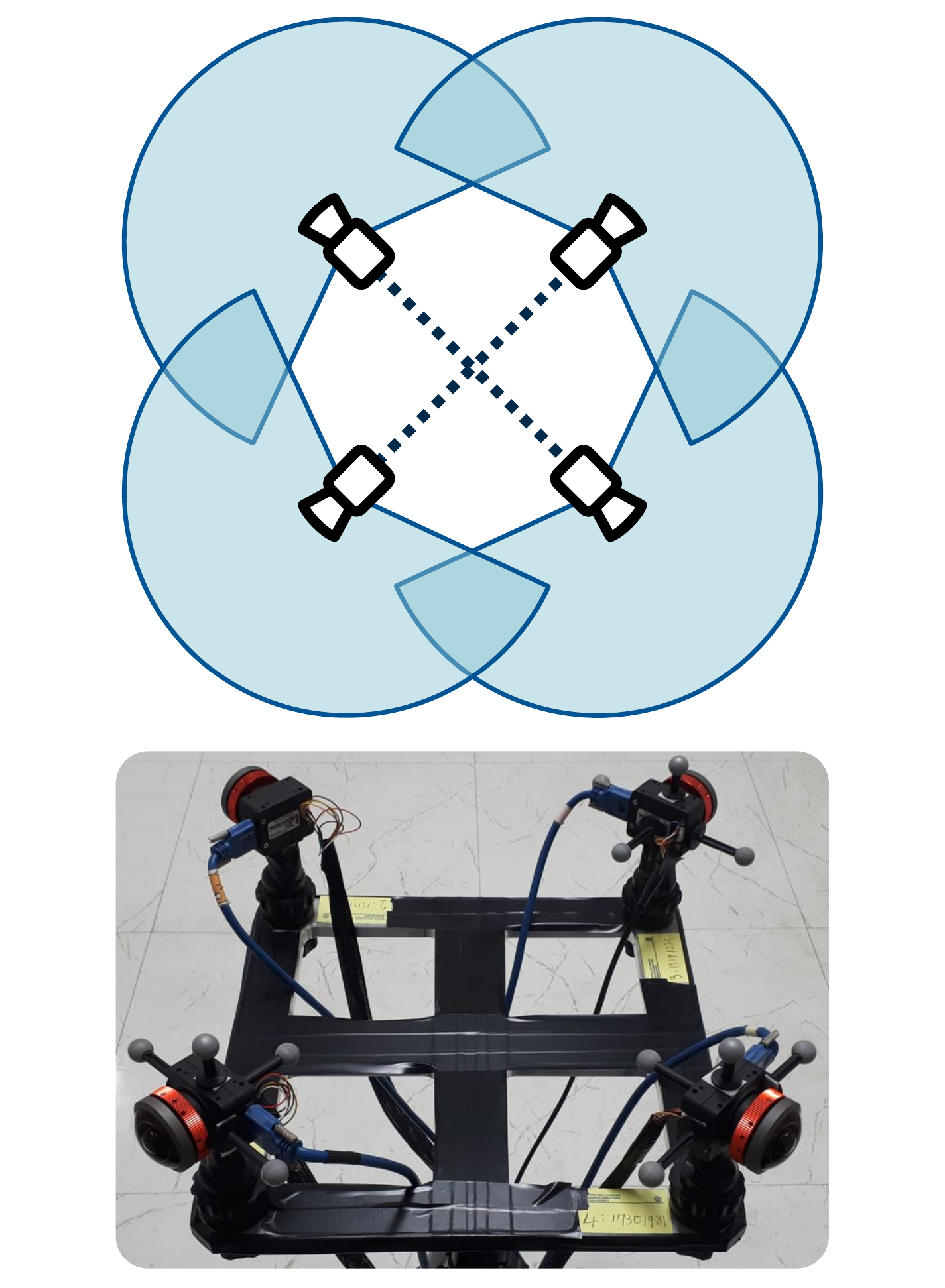}
  \caption{}\label{fig:rig}
  \end{subfigure}
  \begin{subfigure}[b]{0.628\linewidth}
  \centering
  \includegraphics[keepaspectratio=true,width=\linewidth]{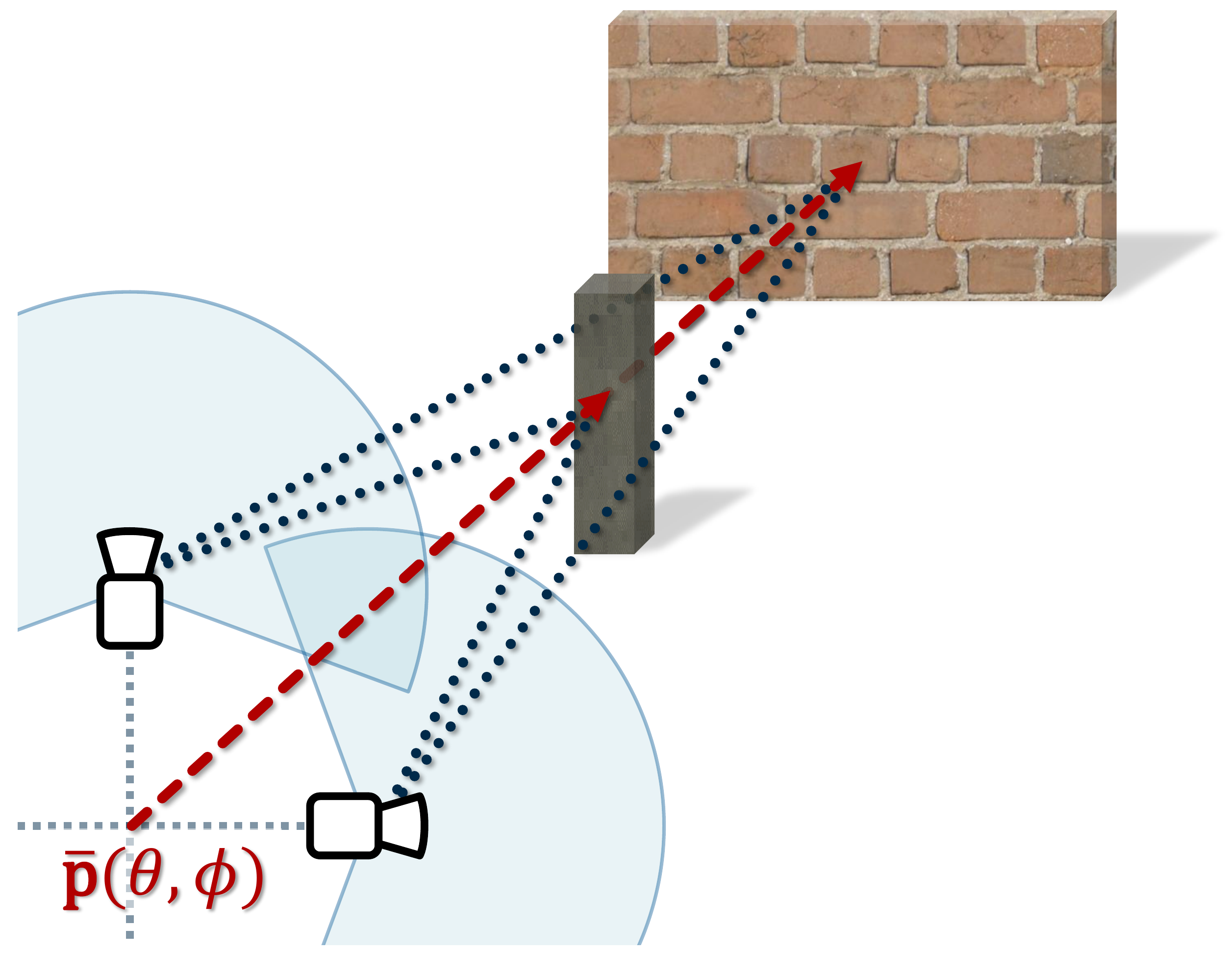}
  \captionsetup{justification=centering}
  \caption{}\label{fig:multitrue}
  \end{subfigure}
  \caption{
(a) \tb{Wide-baseline multi-camera rig system.} (b) \tb{Multiple true matches problem.} There can be several observations on a ray in such global sweeping approach.}
  \label{fig:rig_system}
  \vspace{-8pt}
\end{figure}

Recently, several works have been proposed for the omnidirectional stereo using multiple cameras~\cite{wang2012stereo}, reflective mirrors~\cite{schonbein2014omnidirectional}, or wide FOV fisheye lenses~\cite{gao2017dual}.
Nevertheless, very few works utilize deep neural networks for the omnidirectional stereo.
In SweepNet~\cite{won2019sweepnet} a convolutional neural network (CNN) is used to compute the matching costs of equirectangular image pairs warped from the ultra-wide FOV images.
The result cost volume is then refined by cost aggregation (\eg, Semi-global matching~\cite{hirschmuller2008stereo}), which is a commonly used approach in conventional stereo matching~\cite{chen2015deep, zbontar2016stereo, luo2016efficient}.
However, such an approach may not be optimal in the wide-baseline omnidirectional setup since the occlusions are more frequent and heavier, and there can be multiple true matches for one ray (Fig.~\ref{fig:multitrue}).
%
%
%
%
%
%
On the other hand, recent methods for conventional stereo matching such as GC-Net~\cite{kendall2017end} and PSMNet~\cite{chang2018pyramid} employ the end-to-end deep learning without separate cost aggregation, and achieve better performance compared to the traditional pipeline~\cite{zbontar2016stereo, guney2015displets, seki2017sgm}.

\begin{figure*}[ht]
\centering
    \includegraphics[keepaspectratio=true,width=\textwidth]{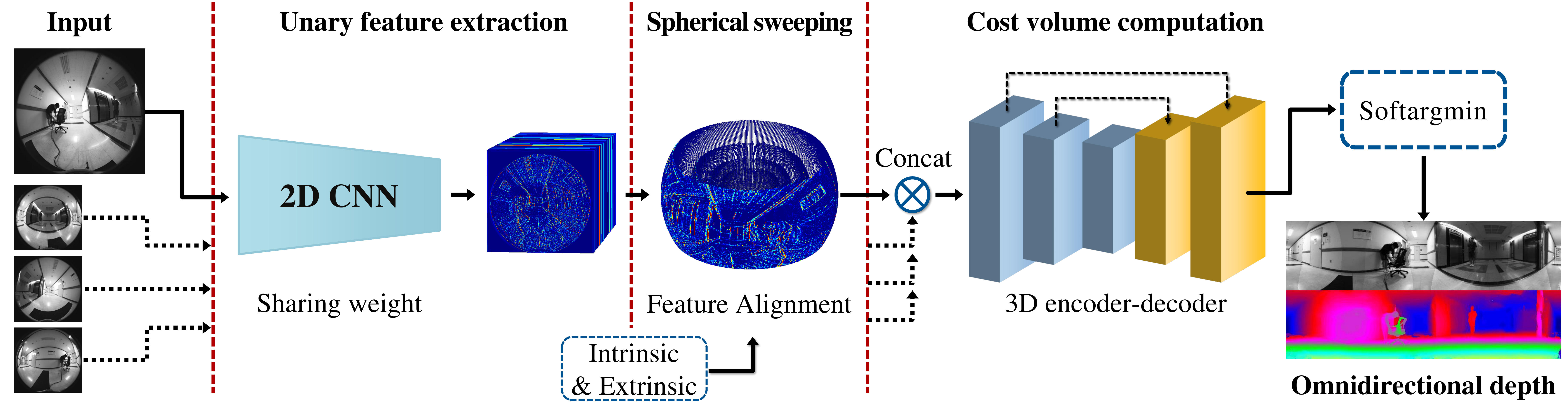}
    \caption{\tb{Overview of the proposed method.} Each input image is fed into the 2D CNN for extracting feature maps. We project the unary feature maps into spherical features to build the matching cost volume. The final depth is acquired through cost volume computation by the 3D encoder-decoder architecture and softargmin.}
    \label{fig:workflow}
    \vspace{-8pt}
\end{figure*}

We introduce a novel end-to-end deep neural network for estimating omnidirectional depth from multi-view fisheye images.
It consists of three blocks, unary feature extraction, spherical sweeping, and cost volume computation as illustrated in Fig.~\ref{fig:workflow}. 
The deep features built from the input images are warped to spherical feature maps for all hypothesized depths (spherical sweeping).
Then a 4D feature volume is formed by concatenating the spherical feature maps from all views so that the correlation between multiple views can be learned efficiently.
Finally, the 3D encoder-decoder block computes a regularized cost volume in consideration of the global context for omnidirectional depth estimation.
While the proposed algorithm can handle various camera layouts, we choose the rig in Fig.~\ref{fig:rig} because it provides good coverage while it can be easily adopted in the existing vehicles.

%

Large-scale data with sufficient quantity, quality, and diversity are essential to train robust deep neural networks.
Nonetheless, acquiring highly accurate dense depth measurements in real-world is very difficult due to the limitations of available depth sensors.
Recent works~\cite{mayer2016large, ros2016synthia} have proposed to use realistically rendered synthetic images with ground truth depth maps for conventional stereo methods.
%
%
Cityscape synthetic datasets in~\cite{won2019sweepnet} are the only available datasets for the omnidirectional multi-view setup, but the number of data is not enough to train a large network, and they are limited to the outdoor driving scenes with few objects.
In this work, we present complementary large-scale synthetic datasets in both indoor and outdoor environments with various objects. 
%

The contributions of this paper are summarized as:
\begin{enumerate}[label=(\roman*)]
\setlength{\topsep}{0pt}
\setlength{\itemsep}{0pt}
\setlength{\partopsep}{0pt} 
\item 
We propose a novel end-to-end deep learning model to estimate an omnidirectional depth from multiple fisheye cameras.
The proposed model directly projects feature maps to the predefined global spheres, combined with the 3D encoder-decoder block enabling to utilize global contexts for computing and regularizing the matching cost.
\item
We offer large-scale synthetic datasets for the omnidirectional depth estimation.
The datasets consist of multiple input fisheye images with corresponding omnidirectional depth maps. 
%
The experiments on the real-world environments show that our datasets successfully train our network.

\item
We experimentally show that the proposed method outperforms the previous multi-stage methods.
We also show that our approaches perform favorably compared to the omnidirectional versions of the state-of-the-art conventional stereo methods through extensive experiments.
%
%
\end{enumerate}

\section{Related Work}

\subsec{Deep Learning-based Methods for Conventional Stereo}
Conventional stereo setup assumes a rectified image pair as the input.
Most traditional stereo algorithms before deep learning follow two steps: matching cost computation and cost aggregation.
As summarized in Hirschmuller~\etal~\cite{hirschmuller2007evaluation}, sum of absolute differences, filter-based cost, mutual information, or normalized cross-correlation are used to compute the matching cost, and for cost aggregation, local correlation-based methods, global graph cuts~\cite{a22001fast}, and semi-global matching (SGM)~\cite{hirschmuller2008stereo} are used.
Among them, SGM~\cite{hirschmuller2008stereo} is widely used because of its high accuracy and low computational overhead.

Recently, deep learning approaches report much improved performance in the stereo matching.
Zagoruyko~\etal~\cite{zagoruyko2015learning} propose a CNN-based similarity measurement for image patch pairs.
Similarly, Zbontar and LeCun~\cite{zbontar2016stereo} introduce MC-CNN that computes matching costs from small image patch pairs.
Meanwhile, several papers focus on the cost aggregation or disparity refinement.
G\"uney and Geiger~\cite{guney2015displets} introduce Displets resolving matching ambiguities on reflection or textureless surfaces using objects' 3D models.
Seki and Pollefeys~\cite{seki2017sgm} propose SGM-Net which predicts the smoothness penalties in SGM~\cite{hirschmuller2008stereo}.

On the other hand, there have been several works on end-to-end modeling of the stereo pipeline.
%
Kendall~\etal~\cite{kendall2017end} propose GC-Net which regularizes the matching cost by 3D convolutional encoder-decoder architecture, and performs disparity regression by the softargmin.
Further, PSMNet by Chang and Chen~\cite{chang2018pyramid} consists of spatial pyramid pooling modules for larger receptive field and multiply stacked 3D encoder-decoder architecture for learning more context information.
Also, Mayer~\etal~\cite{mayer2016large} develop DispNet, an end-to-end network using correlation layers for disparity estimation, and it is further extended by Pang~\etal~\cite{pang2017cascade} (CRL) and Ilg~\etal~\cite{ilg2018occlusions} (DispNet-CSS).
These end-to-end networks have achieved better performance compared to the conventional multi-stage methods.

\subsec{Synthetic Datasets for Learning Stereo Matching}
For successful training of deep neural networks, an adequate large-scale dataset is essential.
%
%
In stereo depth estimation, Middlebury ~\cite{scharstein2003high, hirschmuller2007evaluation, scharstein2014high} and KITTI datasets~\cite{geiger2012we,menze2015object} are most widely used.
These databases are faithfully reflecting the real world, but capturing the ground truth depth requires complex calibration and has limited coverage, and more importantly, the number of images is often insufficient for training large networks.

Nowadays synthetically rendered datasets are used to complement the real datasets.
Mayer~\etal~\cite{mayer2016large} introduce a large scale dataset for disparity, optical flow, and scene flow estimation. 
The proposed dataset consists of 2K scene images and dense disparity maps generated via rendering, which is $10\times$ larger than KITTI~\cite{menze2015object}. 
Ritcher~\etal~\cite{richter2017playing} provide fully annotated training data by simulating a living city in a realistic 3D game world.
For semantic scene completion, SUNGC dataset~\cite{song2016ssc} contains 45K synthetic indoor scenes of 400K rooms and 5M objects with depth and voxel maps.
However, almost all datasets use single or stereo pinhole camera models with limited FOV, and there are very few datasets for omnidirectional stereo.

\subsec{Omnidirectional Depth Estimation}
Various algorithms and systems have been proposed for the omnidirectional depth estimation~\cite{gao2017dual,schonbein2014omnidirectional,wang2012stereo}, but very few use deep neural networks.
Sch\"onbein~\etal~\cite{schonbein2014omnidirectional} use two horizontally mounted $360\degree$-FOV catadioptric cameras, and estimate the disparity from rectified omnidirectional images.
Using two vertically mounted ultra-wide FOV fisheye cameras, Gao and Shen~\cite{gao2017dual} estimate omnidirectional depth by projecting the input images onto four predefined planes.
Im~\etal~\cite{im2016all} propose a temporal stereo algorithm that estimates an all around depth of the static scene from a short motion clip.
Meanwhile, purely learning-based approaches Zioulis~\etal~\cite{zioulis2018omnidepth} and Payen~\etal~\cite{payen2018eliminating} have been proposed estimating a $360\degree$ depth from a single panoramic image. 

Recently, Won~\etal~\cite{won2019sweepnet} propose SweepNet with a multi-camera rig system for the omnidirectional stereo.
They warp the input fisheye images onto the concentric global spheres, and SweepNet computes matching costs from the warped spherical images pair.
Then, the cost volume is refined by applying SGM~\cite{hirschmuller2008stereo}.
However, SGM cannot handle the multiple true matches occurring in such global sweeping approaches as in Fig.~\ref{fig:multitrue}.

In this paper, we present the first end-to-end deep neural network for the omnidirectional stereo and large-scale datasets to train the network.
As shown in the experiments, the proposed method achieves better performance compared to the previous methods and performs favorably in the real-world environment with our new datasets.

\section{Omnidirectional Multi-view Stereo} 

In this section, we introduce the multi-fisheye camera rig and the spherical sweeping method, and then describe the proposed end-to-end network architecture for the omnidirectional stereo depth estimation.
As shown in Fig.~\ref{fig:workflow} our algorithm has three stages, unary feature extraction, spherical sweeping, and cost volume computation.
In the following subsections, the individual stages are described in detail.

\subsection{Spherical sweeping}
The rig consists of multiple fisheye cameras mounted at fixed locations.
Unlike the conventional stereo which uses the reference camera's coordinate system, we use the rig coordinate system for depth representation, as in~\cite{won2019sweepnet}.
For convenience we set the y-axis to be perpendicular to the plane closest to all camera centers, and the origin at the center of the projected camera centers.
A unit ray $\bar{\mathbf{p}}$ for the spherical coordinate $\langle\theta,\phi\rangle$ corresponds to $\bar{\mathbf{p}}(\theta,\phi) = (\cos(\phi)\cos(\theta),\sin(\phi),\cos(\phi)\sin(\theta))^{\top}$.
With the intrinsic and extrinsic parameters of the $i$-th camera (calibrated using \cite{scaramuzza2006flexible, urban2015improved, ceres-solver}), the image pixel coordinate $\mathbf{x}_i$ for a 3D point $\mathbf{X}$ can be written as a projection function $\Pi_i$; $\mathbf{x}_i = \Pi_i(\mathbf{X})$.
Thus a point at $\langle\theta,\phi\rangle$ on the sphere of radius $\rho$ is projected to $\Pi_i(\rho\,\bar{\mathbf{p}}(\theta,\phi))$ in the $i$-th fisheye image.

Spherical sweeping generates a series of spheres with different radii and builds the spherical images of each input image.
Similar to plane sweeping in conventional stereo, the inverse radius $d_n$ is swept from 0 to $d_{max}$, where $1/d_{max}$ is the minimum depth to be considered and $N$ is the number of spheres.
The pixel value of the equirectangular spherical image warped onto $n$-th sphere is determined as
\begin{equation}
    S_{n,i}(\theta,\phi)=I_i(\Pi_i(\bar{\mb{p}}(\theta,\phi)/d_n)),
    \label{eq:spherical}
\end{equation}
where $I_i$ is the input fisheye image captured by $i$-th camera and $d_n$ is the $n$-th inverse depth.

\subsection{Feature Learning and Alignment}
\label{sub:feature}

Instead of using pixel intensities, recent stereo algorithms use deep features for computing matching costs.
%
MC-CNN~\cite{zbontar2016stereo} shifts the right features by $-k$ pixels to align them with the left features, so as to compute the cost for $k$ disparity by $1 \times 1$ convolutional filters.
Further, GC-Net~\cite{kendall2017end} builds a 4D cost volume by shifting and concatenating the feature maps across each disparity, so that it can be regularized by a 3D CNN.
In this way, the network can utilize geometric context (e.g., for handling occlusion) by depth-wise convolution, and also, the simple shifting operation makes gradient back-propagation easy.
However, these approaches are limited to rectified conventional stereo, and cannot be applied to multi-view images in wide FOV or omnidirectional setups.

Instead of extracting features from the spherical images at all spheres, we choose to build a feature map in the input fisheye image space and warp the feature map according to Eq.~\ref{eq:spherical}.
This saves huge amount of computation, and the impact on performance is minimal since the distortion in the original image is learned by the feature extraction network.
The unary feature map $U=F_{CNN}(I)$ has $\frac{1}{r}H_I \times \frac{1}{r}W_I \times C$ resolution, where $F_{CNN}$ is a 2D CNN for the feature extraction, $H_I$ and $W_I$ are the height and width of the input image, $r$ is the reduction factor, and $C$ is the number of channels.

The unary feature maps of the input images are then projected onto the predefined spheres.
Following Eq.~\ref{eq:spherical}, the spherical feature map at $n$-th sphere for $i$-th image is determined as
\begin{equation}
    \mathcal{S}_i(\phi,\theta,n,c) = U_c\left(\frac{1}{r}\Pi_i(\bar{\mb{p}}(\theta,\phi)/d_{n})\right),
\end{equation}
where $\theta$ varies from $-\pi$ to $\pi$, and $\phi$ varies up to $\pm\pi/2$ according to the resolution.
To ensure sufficient disparities between neighboring warped feature maps and to reduce the memory and computation overhead, we use every other spheres, i.e., $n\in[0,2,\ldots,N-1]$, to make the warped 4D feature volume $\mathcal{S}_i$ of the size $H \times W \times \frac{N}{2} \times C$.
With the calibrated intrinsic and extrinsic parameters, we use the coordinate lookup table and 2D bilinear interpolation in warping the feature maps, and during back-propagation the gradients are distributed inversely.
We compute the validity mask $M_i$ for each input image, and the pixels outside the valid region are ignored both in warping and back-propagation.

%
Finally, all spherical feature volumes $\{ \mathcal{S}_i \}$ are merged and used as the input of the cost computation network.
%
%
Our approaches enables the network to learn finding omnidirectional stereo correspondences from multiple fisheye images, and to utilize spherical geometric context information for the regularization by applying a 3D CNN to the spherical features.

\begin{table}[ht!]
\footnotesize
\centering
\resizebox{\linewidth}{!}{
\begin{tabular}{clll} \toprule
& Name & Layer Property & Output $\left(H,W,N,C\right)$ \\ \midrule 
\multirow{7}{*}{\rotatebox[origin=c]{90}{Unary feature}\rotatebox[origin=c]{90}{extraction}}
  & Input &  & $H_{I} \times W_{I}$ \\
& conv1 & $5\times5, 32$ & 
\multirow{6}{*}{ $\left.\vphantom{\begin{tabular}{c}.\\.\\.\\.\\.\\.\end{tabular}}\!\! \right\}$ $\sfrac{1}{2}H_I \times \sfrac{1}{2}W_I \times 32$ } \\
& conv2 & $3\times3, 32$ &  \\
& conv3 & $3\times3, 32$, add conv1 &  \\
& conv4-11 & repeat conv2-3 &  \\
& \multirow{2}{*}{conv12-17} & repeat conv2-3 &  \\ 
&  & with dilate = 2, 3, 4 & \\ \midrule 
\multirow{4}{*}{\rotatebox[origin=c]{90}{Spherical}\rotatebox[origin=c]{90}{sweeping}}  & warp &  & $H \hfill \times \hfill W \hfill \times \hfill \sfrac{1}{2}N \hfill \times \hfill \scriptstyle 32$ \\
& transference & $ 3\times3\times1, 32$ & $\sfrac{1}{2} \hfill \times \sfrac{1}{2} \hfill \times \sfrac{1}{2} \hfill \times \hfill \scriptstyle 32$ \\ \cmidrule{2-4}
& concat(4)* &  & $\sfrac{1}{2} \hfill \times \sfrac{1}{2} \hfill \times \sfrac{1}{2} \hfill \times \scriptstyle 128$ \\
& fusion & $3\times3\times3, 64$ & $\sfrac{1}{2} \hfill \times \sfrac{1}{2} \hfill \times \sfrac{1}{2} \hfill \times \hfill \scriptstyle 64$ \\ \midrule 
\multirow{14}{*}{\rotatebox[origin=c]{90}{Cost volume computation}} 
 & 3Dconv1-3 & $3\times3\times3, 64$ & $\sfrac{1}{2} \hfill \times \sfrac{1}{2} \hfill \times \sfrac{1}{2} \hfill \times \hfill \scriptstyle 64$ \\
& 3Dconv4-6 & from 1, $3\times 3\times 3, 128$ & $\sfrac{1}{4} \hfill \times \sfrac{1}{4} \hfill \times \sfrac{1}{4} \hfill \times \scriptstyle 128$ \\
& 3Dconv7-9 & from 4, $3\times 3\times 3, 128$ & $\sfrac{1}{8} \hfill \times \sfrac{1}{8} \hfill \times \sfrac{1}{8} \hfill \times \scriptstyle 128$ \\
& 3Dconv10-12 & from 7, $3\times 3\times 3, 128$ & $\sfrac{1}{16} \hfill \times \sfrac{1}{16} \hfill \times \sfrac{1}{16} \hfill \times \scriptstyle 128$ \\
& 3Dconv13-15 & from 10, $3\times 3\times 3, 256$ & $\sfrac{1}{32} \hfill \times \sfrac{1}{32} \hfill \times \sfrac{1}{32} \hfill \times \scriptstyle 256$ \\
& 3Ddeconv1 & \begin{tabular}[l]{@{}l@{}}$3\times3\times3, 128,$\\ add 3Dconv12\end{tabular} & $\sfrac{1}{16} \hfill \times  \sfrac{1}{16} \hfill \times  \sfrac{1}{16} \hfill \times \scriptstyle 128$ \\
& 3Ddeconv2 & \begin{tabular}[l]{@{}l@{}}$3\times3\times3, 128,$\\ add 3Dconv9\end{tabular} & $\sfrac{1}{8} \hfill \times  \sfrac{1}{8} \hfill \times \sfrac{1}{8} \hfill \times \scriptstyle 128$ \\
& 3Ddeconv3 & \begin{tabular}[l]{@{}l@{}}$3\times3\times3, 128,$\\ add 3Dconv6\end{tabular} & $\sfrac{1}{4} \hfill \times  \sfrac{1}{4} \hfill \times \sfrac{1}{4} \hfill \times \scriptstyle 128$ \\
& 3Ddeconv4 & \begin{tabular}[l]{@{}l@{}}$3\times3\times3, 64,$\\ add 3Dconv3\end{tabular} & $\sfrac{1}{2} \hfill \times  \sfrac{1}{2} \hfill \times \sfrac{1}{2} \hfill \times \hfill \scriptstyle 64$ \\
& 3Ddeconv5 & $3\times3\times3, 1$ & $H \times W \times N $ \\ \midrule 
& softargmin &  & $H \times W$ \\ \bottomrule
\end{tabular}}
\caption{
The input images pass separately from $\operatorname{conv1}$ to $\operatorname{transference}$, then are merged by $\operatorname{concat}$ and $\operatorname{fusion}$. 
$H$, $W$, and $N$ are omitted for brevity.
In this work we use 4 cameras, thus $\operatorname{concat}$ outputs $32\times 4=128$ channels.
}
\label{tab:network}
\vspace{-8pt}
\end{table}

\subsection{Network Architecture}
\label{sub:architecture}
The architecture of the proposed network is detailed in Table~\ref{tab:network}.
The input of the network is a set of grayscale fisheye images.
We use the residual blocks~\cite{he2016deep} for the unary feature extraction, and the dilated convolution for the larger receptive field.
The output feature map size is half ($r=2$) of the input image.
Each feature map is aligned by the spherical sweeping (Sec.~\ref{sub:feature}), and transferred to the spherical feature by a $3\times3$ convolution.
The spherical feature maps are concatenated and fused into the 4D initial cost volume by a $3\times3\times3$ convolution.
We then use the 3D encoder-decoder architecture~\cite{kendall2017end} to refine and regularize the cost volume using the global context information.

Finally, the inverse depth index $\hat{n}$ can be computed by the softargmin~\cite{kendall2017end} as
\begin{equation*}
    \hat{n}(\theta,\phi)= \sum_{n=0}^{N-1} n \times \frac{e^{-\mathcal{C}(\phi,\theta,n)}}{\sum_{\nu} e^{-\mathcal{C}(\phi,\theta,\nu)}}
\end{equation*}
where $\mathcal{C}$ is the ($H\times W\times N$) regularized cost volume.

To train the network in an end-to-end fashion,
we use the input images and the ground truth inverse depth index as 
\[
n^*(\theta,\phi) = (N-1)\frac{d^*(\theta,\phi) - d_{0}}{d_{N-1}-d_{0}},
\] 
where $d^*(\cdot) = 1/\mathcal{D}^*(\cdot)$ is the ground truth inverse depth, and $d_0$ and $d_{N-1}$ are the min and max inverse depth respectively.
We use the absolute error loss between the ground truth and predicted index as
\begin{equation*}
    L(\theta,\phi) = \frac{1}{\sum_{i} M_i(\theta,\phi)} \Big|\, \hat{n}(\theta,\phi)-\operatorname{round}(n^*(\theta,\phi)) \,\Big|.
\end{equation*}
We use the stochastic gradient descent with a momentum to minimize the loss.
The overall flow of the proposed network is illustrated in Fig.~\ref{fig:workflow}.

\vspace{-10pt}
\section{Datasets}

Although there exist many datasets for conventional stereo~\cite{geiger2012we,menze2015object,mayer2016large}, only one dataset~\cite{won2019sweepnet} is available for the omnidirectional stereo, but it only contains the outdoor road scenes.
Therefore we create new synthetic datasets for more generic scenes and objects.
Our datasets contain input fisheye images, omnidirectional depth maps, and reference panorama images.
%
In addition to \cite{won2019sweepnet}, we generate two much larger datasets (OmniThings and OmniHouse) in different environments using Blender.

\begin{figure}[tb!]
\centering
    \includegraphics[keepaspectratio=true,width=\linewidth]{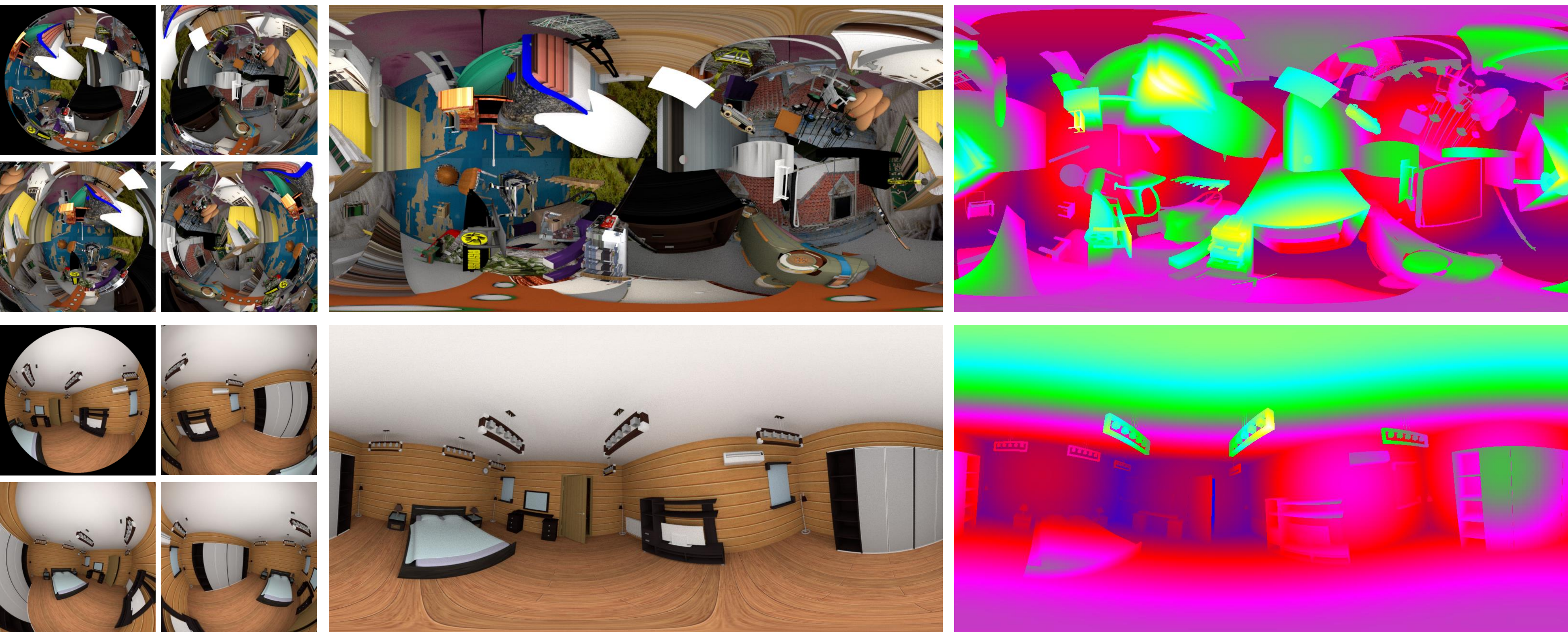}
    \vspace{-8pt}
    \caption{\tb{Examples of our proposed datasets.} From left: input fisheye images with visibility (left-top), reference panorama image, and ground truth inverse depth map.}
    \label{fig:dataset}
\vspace{-4pt}
\end{figure}

\begin{table}[tb!]
\scriptsize
\resizebox{\linewidth}{!}{%
\begin{tabular}{clrrr} \bottomrule
\multicolumn{2}{c}{\multirow{2}{*}{Dataset}} & \multicolumn{1}{c}{\# Training} & \multicolumn{1}{c}{\# Training} & \multicolumn{1}{c}{\# Test}\\
\multicolumn{2}{c}{} & \multicolumn{1}{c}{Scenes} & \multicolumn{1}{c}{Frames} & \multicolumn{1}{c}{Frames} \\ \hline \hline
\multirow{3}{*}{SceneFlow~\cite{mayer2016large}} & FlyingThings3D & 2247 & 21818 & 4248 \\
 & Monkaa & 8 & 8591 & - \\
 & Driving & 1 & 4392 & - \\ \hline
\multirow{3}{*}{Won~\etal~\cite{won2019sweepnet}} & Sunny & 1 & 700 & 300 \\
 & Cloudy & 1 & 700 & 300 \\
 & Sunset & 1 & 700 & 300 \\ \hline
\multirow{2}{*}{Ours} & OmniThings & 9216 & 9216 & 1024 \\
 & OmniHouse & 451 & 2048 & 512 \\ \toprule
\end{tabular}
}
\vspace{-4pt}
\caption{\tb{Comparison with the published datasets.} Our datasets have more training scenes and the comparable number of training and test frames to exsiting datasets.}
\label{tab:db}
\vspace{-8pt}
\end{table}

\subsec{OmniThings}
Similar to~\cite{mayer2016large}, OmniThings dataset consists of randomly generated objects around the camera rig.
We collect 33474 3D object models from ShapeNet~\cite{chang2015shapenet} and 4711 textures from Flickr and ImageAfter\footnote{https://www.flickr.com and http://www.imageafter.com}.
For each scene, we randomly choose 64 objects and place them onto the $N$ spheres with random positions, rotations, scales, and textures, so that complex shapes of various objects and occlusions can be learned.
We also place a randomly shaped room or sky for learning the background depth.
The dataset has 9216 scenes for training and 1024 scenes for test.

\subsec{OmniHouse}
In order to generate realistic indoor scenes, we reproduce the SUNCG dataset~\cite{song2016ssc} which consists of 45K synthetic indoor scenes. 
We collect 451 house models from the SUNCG dataset, and place the virtual camera rig in them with random positions and orientations.
We render 2048 frames for training and 512 frames for test.

The overview of our proposed datasets is described in Table~\ref{tab:db}, and the examples are shown in Fig.~\ref{fig:dataset}.
Each frame consists of four $220\degree$ FOV fisheye images, which have a resolution of $H_I=768$ and $W_I=800$, and one ground truth omnidirectional depth map, which has $H=360$ and $W=640$ ($\theta$ ranges from $-\pi$ to $\pi$ and $\phi$ from $-\pi/2$ to $\pi/2$).
In the next section, we show that the networks trained with our datasets successfully estimate the omnidirectional depth in the real-world environments, which proves the effectiveness of our synthetic datasets.

\vspace{-3pt}
\section{Experimental Results}

\begin{table*}[tb!]
\centering
\resizebox{0.97\textwidth}{!}{%
\begin{tabular}{cl|rrrrrr|rrrrrc} \bottomrule
\multicolumn{2}{l|}{~~Dataset} & \multicolumn{6}{c|}{OmniThings} & \multicolumn{5}{c}{OmniHouse} & \\

\multicolumn{2}{l|}{~~Metric} & \multicolumn{1}{c}{~\textgreater{}1~} & \multicolumn{1}{c}{~\textgreater{}3~} & \multicolumn{1}{c}{~\textgreater{}5~} & \multicolumn{1}{c}{~MAE~} & \multicolumn{1}{c}{~RMS~} & & \multicolumn{1}{c}{~\textgreater{}1~} & \multicolumn{1}{c}{~\textgreater{}3~} & \multicolumn{1}{c}{~\textgreater{}5~} & \multicolumn{1}{c}{MAE} & \multicolumn{1}{c}{~RMS~} &\\ 

\toprule \multicolumn{12}{l}{\tb{~~Spherical sweeping with regularization}} \\ \hline
\multicolumn{2}{l|}{~~ZNCC+SGM~\cite{hirschmuller2008stereo}} &72.56& 54.01 & 45.63 & 10.51 & 16.44 & & 44.05 & 20.64 & 13.57 & 3.08 & 7.05 &\\
\multicolumn{2}{l|}{~~MC-CNN~\cite{zbontar2016stereo}+SGM} & 67.19 & 47.43 & 39.49 & 8.65 & 13.66 & & 38.01 & 15.86 & 9.46 & 2.08 & 4.15 &\\
\multicolumn{2}{l|}{~~SweepNet~\cite{won2019sweepnet}+SGM} & 67.20 & 47.63 & 39.66 & 8.87 & 13.90 & & 36.60 & 15.41 & 9.36 & 2.07 & 4.38 &\\

\toprule \multicolumn{12}{l}{\tb{~Stitching conventional stereo}} \\ \hline
\multicolumn{2}{l|}{~~PSMNet~\cite{chang2018pyramid}} & 86.25 & 63.23 & 44.84 & 7.28 & 11.15 & & 63.22 & 26.43 & 15.39 & 5.82 & 13.88 &\\
\multicolumn{2}{l|}{~~PSMNet-ft} & 82.69 & 51.98 & 41.74 & 9.09 & 13.71 & & 87.56 & 27.01 & 12.89 & 3.51 & 6.05 &\\
\multicolumn{2}{l|}{~~DispNet-CSS~\cite{ilg2018occlusions}} & 50.62 & 27.77 & 19.50 & 4.06 & 7.98 & & $^*$26.56 & 11.69 & 7.16 & $^*$1.54 & $^*$3.18 &\\
\multicolumn{2}{l|}{~~DispNet-CSS-ft} & 67.86 & 48.08 & 38.57 & 7.81 & 12.27 & & 36.47 & 14.98 & 8.29 & 1.81 & 3.44 &\\
\toprule \hline

\multicolumn{2}{l|}{~~\tb{OmniMVS}} & \tb{47.72} & \tb{15.12} & \tb{8.91} & \tb{2.40} & \tb{5.27} & & 30.53 & $^*$10.29 & $^*$6.27 & 1.72 & 4.05 &\\
\multicolumn{2}{l|}{~~\tb{OmniMVS-ft}} & $^*$50.28 & $^*$22.78 & $^*$15.60 & $^*$3.52 & $^*$7.44 & & \textbf{21.09} & \textbf{4.63} & \textbf{2.58} & \textbf{1.04} & \textbf{1.97}& \\ \toprule
\end{tabular}
}
\vspace{-3pt}
\resizebox{0.97\textwidth}{!}{%
\begin{tabular}{l|rrrrr|rrrrr|rrrrr} \bottomrule
\multicolumn{1}{l|}{Dataset} & \multicolumn{5}{c|}{Sunny} & \multicolumn{5}{c|}{Cloudy} & \multicolumn{5}{c}{Sunset} \\
\multicolumn{1}{l|}{Metric} & \multicolumn{1}{c}{\textgreater{}1} & \multicolumn{1}{c}{\textgreater{}3} & \multicolumn{1}{c}{\textgreater{}5} & \multicolumn{1}{c}{MAE} & \multicolumn{1}{c|}{RMS} & \multicolumn{1}{c}{\textgreater{}1} & \multicolumn{1}{c}{\textgreater{}3} & \multicolumn{1}{c}{\textgreater{}5} & \multicolumn{1}{c}{MAE} & \multicolumn{1}{c|}{RMS} & \multicolumn{1}{c}{\textgreater{}1} & \multicolumn{1}{c}{\textgreater{}3} & \multicolumn{1}{c}{\textgreater{}5} & \multicolumn{1}{c}{MAE} & \multicolumn{1}{c}{RMS} \\ 

\toprule \multicolumn{16}{l}{\tb{Spherical sweeping with regularization}} \\ \hline
ZNCC+SGM & 52.00 & 21.45 & 10.96 & 2.50 & 5.35 & 53.09 & 22.17 & 11.50 & 2.58 & 5.45 & 52.33 & 21.90 & 11.29 & 2.53 & 5.31 \\
MC-CNN+SGM & 39.42 & 11.73 & 6.08 & 1.83 & 4.56 & 43.16 & 11.95 & 5.82 & 1.85 & 4.46 & 39.67 & 12.82 & 6.28 & 1.86 & 4.59 \\
SweepNet+SGM& 24.82 & 6.91 & 4.28 & 1.31 & 3.79 & 34.97 & 9.51 & 5.09 & 1.55 & 3.96 & 24.92 & 7.25 & 4.46 & 1.32 & 3.80 \\

\toprule \multicolumn{16}{l}{\tb{Stitching conventional stereo}} \\ \hline
PSMNet & 65.09 & 30.87 & 13.13 & 2.54 & 4.03 & 63.62 & 28.51 & 10.40 & 2.45 & 4.26 & 63.83 & 28.41 & 10.00 & 2.43 & 4.11 \\
PSMNet-ft & 92.67 & 31.45 & 21.32 & 4.33 & 7.76 & 92.92 & 31.24 & 20.14 & 4.13 & 7.32 & 93.24 & 30.64 & 19.65 & 4.11 & 7.43 \\
DispNet-CSS & $^*$24.80 & 8.54 & 5.59 & 1.44 & 4.02 & $^*$25.16 & 8.47 & 5.50 & 1.43 & 3.92 & $^*$24.79 & 8.29 & 5.34 & 1.38 & 3.76\\
DispNet-CSS-ft & 39.02 & 21.12 & 14.47 & 2.37 & 4.85 & 42.29 & 21.55 & 14.28 & 2.43 & 4.88 & 40.21 & 20.91 & 14.43 & 2.40 & 4.88 \\
\toprule \hline

\tb{OmniMVS} & 27.16 & $^*$6.13 & $^*$3.98 & $^*$1.24 & $^*$3.09 & 28.13 & $^*$5.37 & $^*$3.54 & $^*$1.17 & $^*$2.83 & 26.70 & $^*$6.19 & $^*$4.02 & $^*$1.24 & $^*$3.06\\
\tb{OmniMVS-ft} & \tb{13.93} & \tb{2.87} & \tb{1.71} & \tb{0.79} & \tb{2.12} & \tb{12.20} & \tb{2.48} & \tb{1.46} & \tb{0.72} & \tb{1.85} & \tb{14.14} & \tb{2.88} & \tb{1.71} & \tb{0.79} & \tb{2.04} \\ \toprule
\end{tabular}
}


\vspace{-4pt}
\caption{\tb{Quantitative comparison with other methods.} The error is defined in Eq. \ref{eq:error}. The qualifier \textquotesingle\textgreater{n}\textquotesingle~refers to the pixel ratio (\%) whose error is larger than n, \textquotesingle MAE\textquotesingle~refers to the mean absolute error, and \textquotesingle RMS\textquotesingle~refers to the root mean squared error. The errors are averaged over all test frames of each datasets. \textquotesingle$^*$\textquotesingle~of each scores denotes the 2nd place.}
\label{tab:quan}
\vspace{-1em}
\end{table*}

\begin{table}[tb!]
\resizebox{\linewidth}{!}{%
\begin{tabu}{lccccc}\bottomrule
\multirow{2}{*}{Dataset} & \multicolumn{3}{c}{Omnidirectional stereo} & \multicolumn{2}{c}{Conventional stereo} \\ 
 & Sunny~\cite{won2019sweepnet} & \begin{tabular}[c]{@{}c@{}}Omni\\ Things\end{tabular} & \begin{tabular}[c]{@{}c@{}}Omni\\ House\end{tabular} & \begin{tabular}[c]{@{}c@{}}Scene\\ Flow\end{tabular}~\cite{mayer2016large} & KITTI~\cite{geiger2012we, menze2015object} \\ \hline
MC-CNN~\cite{zbontar2016stereo} & \checkmark &  &  &  &  \\
SweepNet~\cite{won2019sweepnet} & \checkmark &  &  &  &  \\
PSMNet~\cite{chang2018pyramid} &  &  &  & \checkmark &  \\
PSMNet-ft &  &  &  & \checkmark & \checkmark \\
DispNet-CSS~\cite{ilg2018occlusions} &  &  &  & \checkmark &  \\
DispNet-CSS-ft &  &  &  & \checkmark & \checkmark  \\ \tabucline[on 1.5pt off 2pt]-
OmniMVS &  & \checkmark &  &  &  \\
OmniMVS-ft & \checkmark & \checkmark & \checkmark &  & \\ \toprule
\end{tabu}
}
\vspace{-8pt}
\caption{\tb{Datasets used in each methods.} For experimental comparisons we use the published pre-trained weights for other methods (up of the dashed line). \textquotesingle-ft\textquotesingle~ denotes the fine-tuned versions.}
\label{tab:train_db}
\vspace{-8pt}
\end{table}

\subsection{Implementation and Training Details}

\begin{figure}[tb!]
\centering
    \includegraphics[keepaspectratio=true,width=\linewidth]{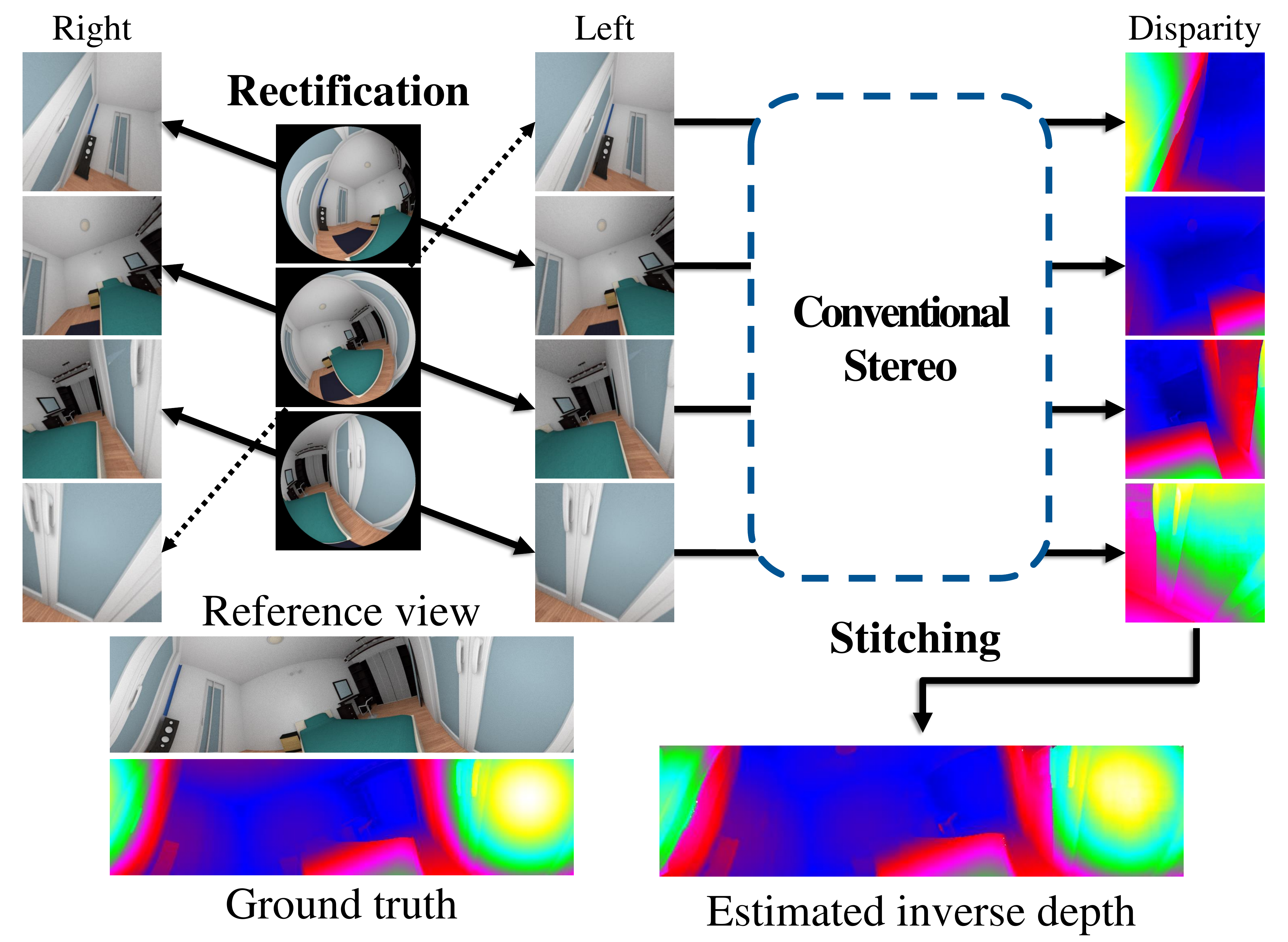}
    \vspace{-4pt}
    \caption{We rectify the input images into $512\times512$ and $120\degree$ FOV left-right pairs. The predicted disparity maps are merged into a $H\times W$ omnidirectional inverse depth index.}
    \label{fig:rectify}
    \vspace{-8pt}
\end{figure}

To train the network, the input images are converted to grayscale, and the validity mask is set to only contain the pixels within  $220\degree$ FOV.
The intensity values in the valid area are then normalized to zero-mean and unit variance.
To prevent the encoder-decoder network from learning only the valid regions in each channel, the order of feature maps to the concatenating stage is randomly permuted (\eg, 1-2-3-4, 2-3-4-1, 3-4-1-2, or 4-1-2-3).
Further, we randomly rotate the rig coordinate system (and the GT depth map accordingly) with a small angle, so that the network is not tightly coupled to specific layouts.
In all our experiments, the output and GT depth maps are cropped to $H=160$ ($-\pi/4 \leq \phi \leq \pi/4)$ and $W=640$  since the regions near the poles are highly distorted and less useful.
The number of sweep spheres is set to $N=192$.
We train our network for 30 epochs on the OmniThings dataset from scratch, using 4096 training scenes.
The learning rate $\lambda$ is set to $0.003$ for the first 20 epochs and $0.0003$ for the remaining 10 epochs.
We also test the network fine-tuned on the Sunny and OmniHouse datasets for 16 epochs, with $\lambda=0.003$ for 12 epochs and $\lambda=0.0003$ for the rest.
In our system with a Nvidia 1080ti, our OmniMVS takes 1.06s for processing which is quite fast, where MC-CNN~\cite{zbontar2016stereo} takes 1.97s, SweepNet~\cite{won2019sweepnet} 6.16s, PSMNet~\cite{chang2018pyramid} 1.79s, and DispNet-CSS~\cite{ilg2018occlusions} 0.57s.

\subsection{Quantitative Evaluation}

The error is measured by tie difference of inverse depth index as
\begin{equation}
\label{eq:error}
    E(\phi,\theta) = \frac{\left| \hat{n}(\phi,\theta) - n^*(\phi,\theta) \right| }{N} \times 100,
\end{equation}
which is the percent error of estimated inverse depth index from GT compared to all possible indices ($N$).
We evaluate our approaches quantitatively on the available omnidirectional stereo datasets (Sunny, Cloudy, Sunset~\cite{won2019sweepnet}, OmniThings and OmniHouse).

We compare our method to the previous works of two types.
The first type is spherical sweeping-based omnidirectional methods, and the second is stitching conventional stereo results into an omnidirectional one.
%
We use the pre-trained weights of other methods in testing, and the training datasets for each method are described in Table~\ref{tab:train_db}. 

\begin{figure*}[ht!]
\centering
    \includegraphics[keepaspectratio=true,width=0.97\textwidth]{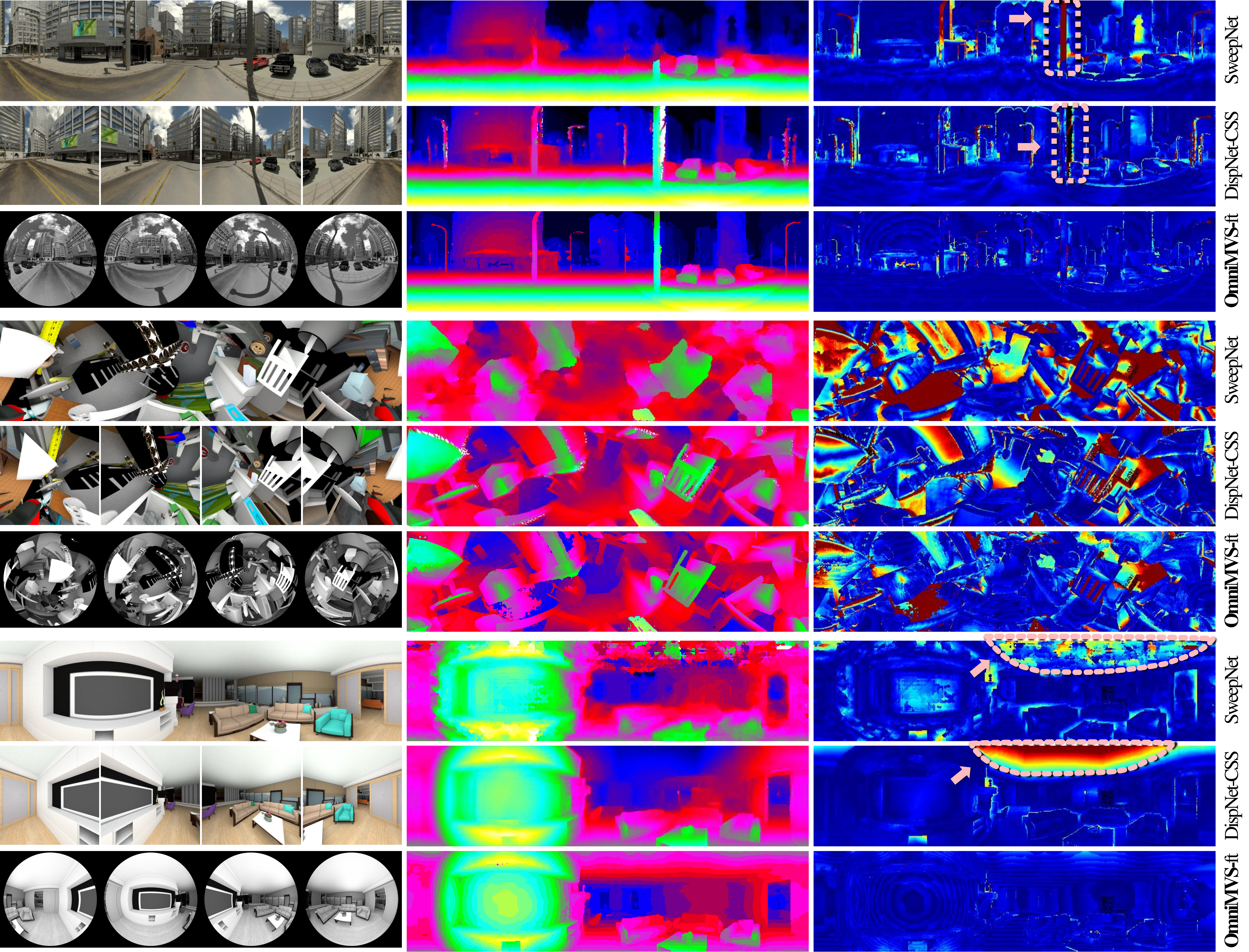}
    \caption{\tb{Results on the synthetic data.} Left: reference panorama image, rectified left color images, and grayscale fisheye images. Middle: predicted inverse depth. Right: colored error map of inverse depth index error (blue is low and red is high). }
    \label{fig:qualitative}
    \vspace{-8pt}
\end{figure*}

\subsec{Spherical sweeping}
%
ZNCC (zero-mean normalized cross correlation) and MC-CNN~\cite{zbontar2016stereo} compute the matching cost from $9\times9$ patches pair in the warped spherical images, and SweepNet~\cite{won2019sweepnet} estimates the whole matching cost volume from the spherical images pair.
%
%
Then, SGM~\cite{hirschmuller2008stereo} regularizes the cost volume with the smoothness penalties $P_1=0.1$ and $P_2=12.0$.
As shown in Table~\ref{tab:quan}, our end-to-end networks perform better in all datasets and metrics.
Our OmniMVS builds more effective feature maps and learns better matching and regularization.

\begin{figure*}[ht!]
\centering
    \includegraphics[keepaspectratio=true,width=0.97\textwidth]{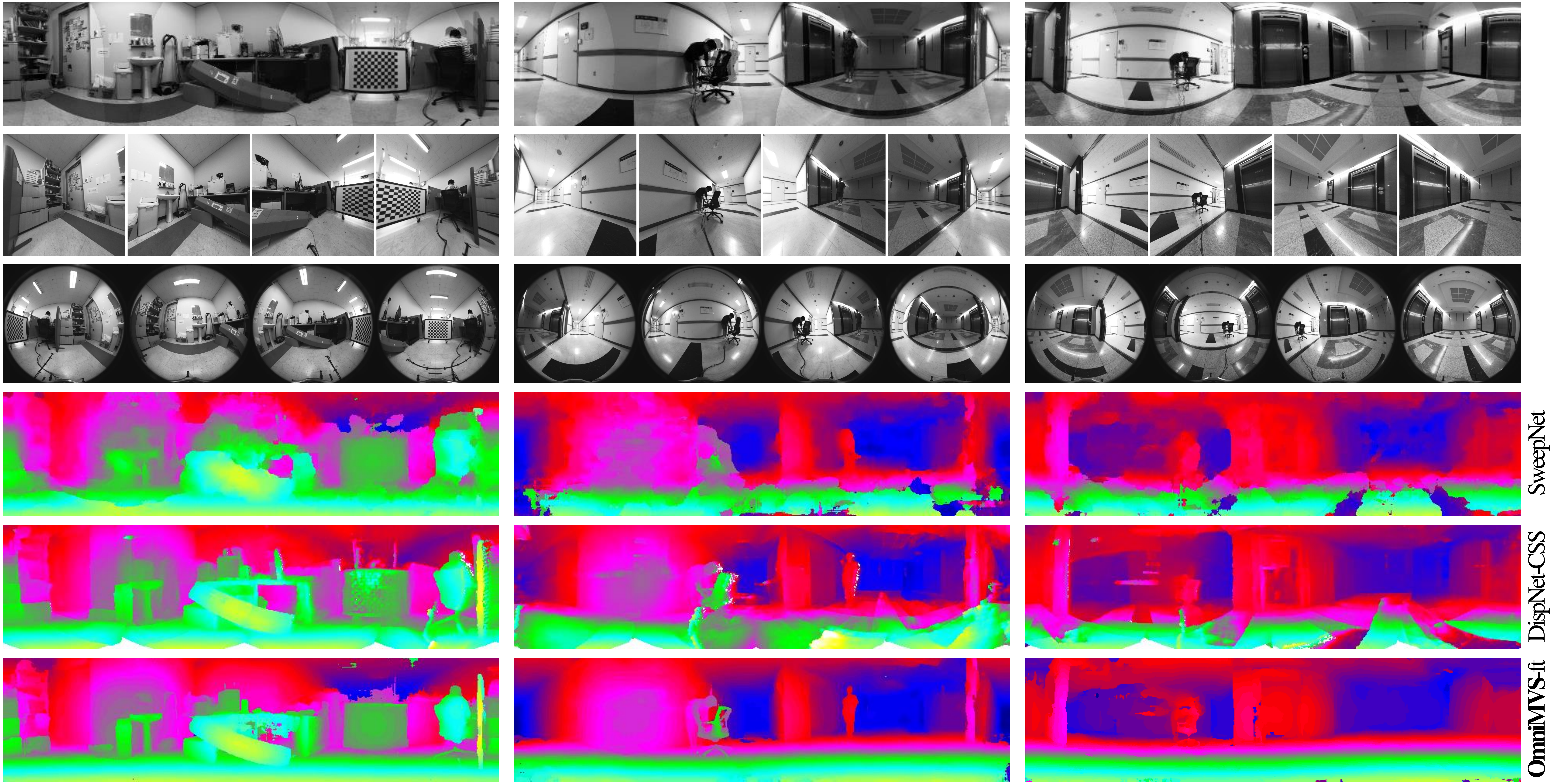}
    \vspace{-2pt}
    \caption{\tb{Results on the real data.} From top: reference panorama image, rectified left images, input grayscale fisheye images, and inverse depth maps predicted by each methods. The reference panorama images are created by projecting the estimated 3D points from OmniMVS-ft to the input images.}
    \label{fig:real}
    \vspace{-8pt}
\end{figure*}

\begin{figure*}[ht!]
\centering
    \includegraphics[keepaspectratio=true,width=0.97\textwidth]{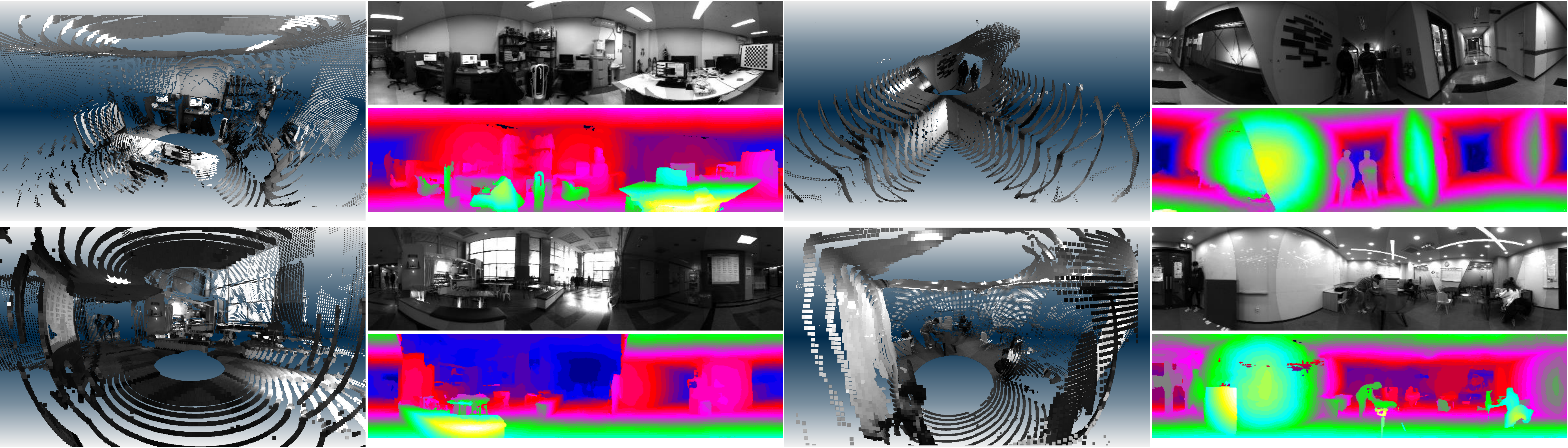}
     \vspace{-2pt}
    \caption{\tb{Point cloud results.} Left: point cloud. Right: reference panorama image and predicted inverse depth estimated by the proposed OmniMVS-ft.
    Note that texureless walls are straight and small objects are reconstructed accurately.
    It also can handle generic rig poses (top-right).
    }
    \label{fig:pointcloud}
\vspace{-1em}
\end{figure*}

\subsec{Stitching Conventional Stereo}
%
In order to estimate an omnidirectional depth, one can use a conventional stereo method to compute disparities in different directions, and merge the depth maps into one panorama.
As shown in Fig.~\ref{fig:rectify}, we generate four $120\degree$ rectified RGB image pairs from the fisheye images, and compute disparities by applying PSMNet\footnote{https://github.com/JiaRenChang/PSMNet}~\cite{chang2018pyramid} or DispNet-CSS\footnote{https://github.com/lmb-freiburg/netdef\_models}~\cite{ilg2018occlusions}.
Then all reconstructed 3D points are put in the rig coordinate system.
For each pixel in the $H\times W$ spherical depth map, the closest 3D point which is projected within 1-pixel radius is chosen for output.
%
%
The pixels without any points are ignored in the evaluation.
As described in Table~\ref{tab:train_db}, we use the pre-trained weights presented in their works.
%
Table~\ref{tab:quan} shows that our networks achieved the best performance.
Note that although OmniThngs and FlyingThings3D in SceneFlow~\cite{mayer2016large} which share most of the objects, OmniMVS trained with OmniThings performs favorably to PSMNet or DispNet-CSS trained with SceneFlow.

\subsection{Qualitative Evaluation}

\subsec{Synthetic Dataset}
Figure~\ref{fig:qualitative} illustrates qualitative results of SweepNet~\cite{won2019sweepnet}, DispNet-CSS~\cite{ilg2018occlusions}, and OmniMVS-ft on the synthetic datasets, Sunny, OmniThings and OmniHouse.
As indicated by the orange arrows in Fig.~\ref{fig:qualitative}, SweepNet with SGM~\cite{hirschmuller2008stereo} does not handle the multiple true matches properly (on the street lamp and background buildings) so the depth of thin objects is overridden by the background depth.
Also they have difficulty in dealing with large textureless regions.
Our network can successfully resolve these problems using global context information.

\subsec{Real-world Data}
We show the capability of our proposed algorithm with real-world data~\cite{won2019sweepnet}.
In all experiments, we use the same configuration with the synthetic case and the identical networks without retraining.
%
%
%
As shown in Fig.~\ref{fig:real} and \ref{fig:pointcloud}, our network generates clean and detailed reconstructions of large textureless or even reflective surfaces as well as small objects like people and chairs.
%

\section{Conclusions}

In this paper we propose a novel end-to-end CNN architecture, OmniMVS for the omnidirectional depth estimation.
%
The proposed network first converts the input fisheye images into the unary feature maps, and builds the 4D feature volume using the calibration and spherical sweeping.
%
The 3D encoder-decoder block computes the matching cost volume, and the final depth estimate is computed by softargmin.
Out network can learn the global context information and successfully reconstructs accurate omnidirectional depth estimates even for thin and small objects as well as large textureless surfaces.
We also present large-scale synthetic datasets, Omnithings and OmniHouse.
The extensive experiments show that our method outperforms existing omnidirectional methods and the state-of-the-art conventional stereo methods with stitching.
\vspace{-5pt}
\small{
\section*{Acknowledgement}
\vspace{-5pt}
This research was supported by Next-Generation Information Computing Development program through National Research Foundation of Korea (NRF) funded by the Ministry of Science, ICT (NRF-2017M3C4A7069369), the NRF grant funded by the Korea government(MSIP)(NRF-2017R1A2B4011928), Research Fellow program funded by the Korea government (NRF-2017R1A6A3A11031193), and Samsung Research Funding \& Incubation Center for Future Technology (SRFC-TC1603-05).
}


{\small
\bibliographystyle{ieee_fullname}
\bibliography{egbib}
}
\clearpage
\includepdf[pages=1]{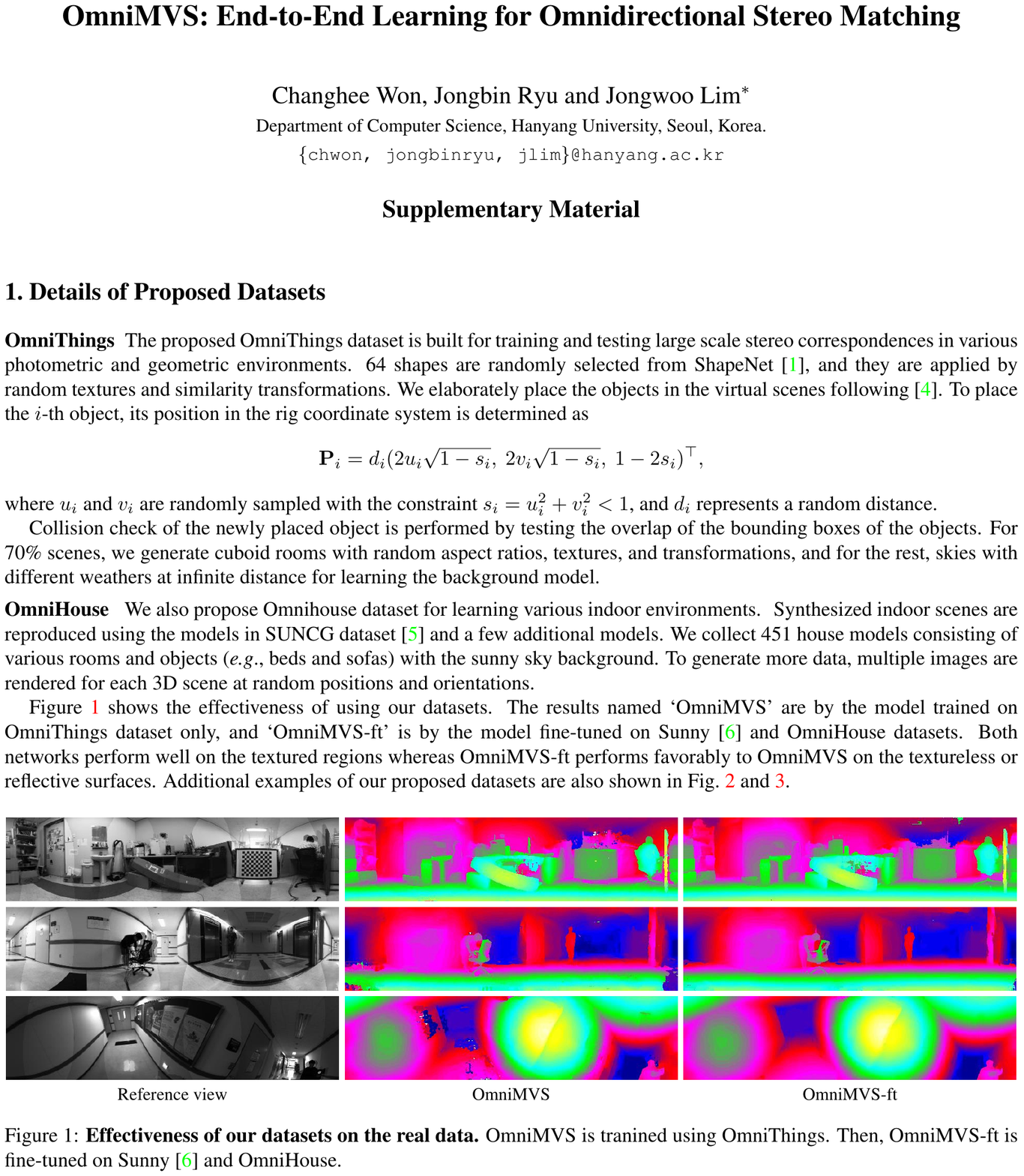}
\includepdf[pages=2]{iccv-supp}
\includepdf[pages=3]{iccv-supp}
\includepdf[pages=4]{iccv-supp}

\end{document}


\title{
OmniMVS: End-to-end Learning for Omnidirectional Stereo Matching
}

\author{
Anonymous ICCV submission \\ 
\\ Paper ID 4237 \\
\\ \tb{Supplementary Material}
}

\maketitle

\pagestyle{plain}
\section{Details of Proposed Datasets}

\subsec{OmniThings}
OmniThings dataset is created with the purpose of learning to find stereo correspondences in various photometric and geometric environments. 
In order to generate scenes, we randomly choose 64 shapes from ShapeNet~\cite{chang2015shapenet} and adapt random textures, rotation, and scale transform to them.
We follow~\cite{marsaglia1972choosing} to efficiently place the objects on everywhere around the virtual camera rig.
For $i$-th object's position vector $\mb{P}_i$, two values $V_{i,1}$ and $V_{i,2}$ are generated randomly in the interval $(-1,1)$ until satisfying $S_i=V_{i,1}^2+V_{i,2}^2 < 1$.
Then, $\mb{P}_i$ is determined as
    $\mb{P}_i = (2V_{i,1}\sqrt{1-S_i},~2V_{i,2}\sqrt{1-S_i},~1-2S_i)^\top$.
%
We displace the objects by multiplying random distances on $\{ \mb{P}_i \}$.
We do not consider whether the objects are overlapping with each other, but we compute each object's bounding box so as to ensure minimum distance between the rig.
We also generate cuboid room with random ratio, scale, and textures, or skies with different weathers for learning the background depth.

\subsec{OmniHouse}
In the real-world environments, there exist many textureless, reflective, or occluded regions which are hard to be matched in the stereo images pair.
To this end, we produce Omnihouse dataset with the purpose of learning to compute and regularize matching costs of the ambiguous regions using the geometric context information.
OmniHouse consists of synthetic indoor scenes reproduced by using the SUNCG dataset~\cite{song2016ssc} (some scenes are collected by ourselves).
We collect 451 house model with various rooms and indoor objects such as people, TV, bed and sofa.
We set the background texture with Sunny skies.
The scenes are captured by the virtual rig placed in the house with random position and orientations.

\begin{figure*}[hbt!]
\centering
    \includegraphics[keepaspectratio=true,width=\textwidth]{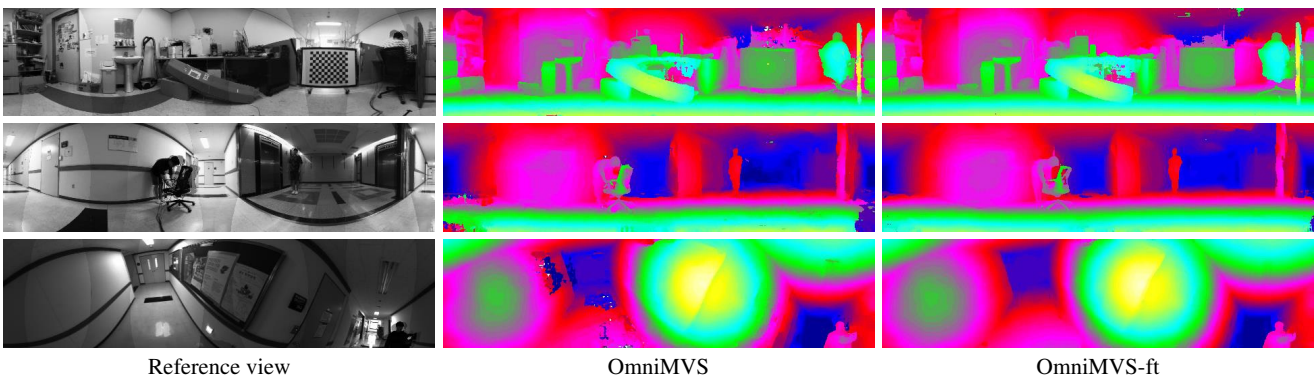}
    \caption{\tb{Effectiveness of our datasets on the real data.} OmniMVS is tranined using OmniThings. Then, OmniMVS-ft is fine-tuned on Sunny~\cite{won2019sweepnet} and OmniHouse.}
    \label{fig:omni-ft}
\end{figure*}

Figure~\ref{fig:omni-ft} shows the effectiveness of our datasets.
OmniMVS is trained on OmniThings dataset and then OmniMVS-ft is fine-tuned on Sunny~\cite{won2019sweepnet} and OmniHouse.
Both networks performs well on the textured regions whereas OmniMVS-ft performs favorably to OmniMVS on the textureless or reflective surfaces.
Additional examples of our proposed datasets are also shown in Fig.~\ref{fig:omnihouse}. 

\section{Grayscale Input for Conventional Stereo}

Altough the conventional stereo methods, PSMNet~\cite{chang2018pyramid} and DispNet-CSS~\cite{ilg2018occlusions}, are trained on the RGB input images,
we use the grayscale images in the experiments on the real data.
We therefore demonstrate that the performance differences between using RGB and grayscale images for the methods are slight (or even better).

\begin{table*}[tb!]
\centering
\resizebox{\textwidth}{!}{%
\begin{tabular}{c|l|rrrrrr|rrrrrc} \bottomrule
\multicolumn{2}{l|}{Dataset} & \multicolumn{6}{c|}{OmniThings} & \multicolumn{5}{c}{OmniHouse} & \\

\multicolumn{2}{l|}{Metric} & \multicolumn{1}{c}{~\textgreater{}1~} & \multicolumn{1}{c}{~\textgreater{}3~} & \multicolumn{1}{c}{~\textgreater{}5~} & \multicolumn{1}{c}{~MAE~} & \multicolumn{1}{c}{~RMS~} & & \multicolumn{1}{c}{~\textgreater{}1~} & \multicolumn{1}{c}{~\textgreater{}3~} & \multicolumn{1}{c}{~\textgreater{}5~} & \multicolumn{1}{c}{MAE} & \multicolumn{1}{c}{~RMS~} &\\ 

\toprule \hline

\multirow{4}{*}{\rotatebox[origin=c]{90}{RGB}} & PSMNet~\cite{chang2018pyramid} & 86.25 & 63.23 & 44.84 & 7.28 & 11.15 & & 63.22 & 26.43 & 15.39 & 5.82 & 13.88 &\\
& PSMNet-ft & 82.69 & 51.98 & 41.74 & 9.09 & 13.71 & & 87.56 & 27.01 & 12.89 & 3.51 & 6.05 &\\
& DispNet-CSS~\cite{ilg2018occlusions} & \tb{50.62} & \tb{27.77} & \tb{19.50} & \tb{4.06} & \tb{7.98} & & \tb{26.56} & \tb{11.69} & \tb{7.16} & \tb{1.54} & \tb{3.18} &\\
& DispNet-CSS-ft & 67.86 & 48.08 & 38.57 & 7.81 & 12.27 & & 36.47 & 14.98 & 8.29 & 1.81 & 3.44 &\\ \hline
\multirow{4}{*}{\rotatebox[origin=c]{90}{Gray}} & PSMNet & 86.77 & 64.33 & 45.78 & 7.42 &  10.72 & & 64.32 & 25.48 & 13.59 & 4.40 & 10.61 &\\
& PSMNet-ft & 82.64 & 51.70 & 41.69 & 9.16 & 13.83 & & 87.70 & 26.90 & 13.12 & 3.56 & 6.18 &\\
& DispNet-CSS & 50.64 & 27.92 & 19.62 & 4.06 & 8.00 & & 26.61 & 11.83 & 7.30 & 1.56 & 3.22 &\\
& DispNet-CSS-ft & 67.90 & 48.24 & 38.76 & 7.89 & 12.37 & & 36.48 & 15.08 & 8.46 & 1.82 & 3.46
 &\\
\toprule
\end{tabular}
}

\resizebox{\textwidth}{!}{%
\begin{tabular}{c|l|rrrrr|rrrrr|rrrrr} \bottomrule
\multicolumn{2}{l|}{Dataset} & \multicolumn{5}{c|}{Sunny} & \multicolumn{5}{c|}{Cloudy} & \multicolumn{5}{c}{Sunset} \\
\multicolumn{2}{l|}{Metric} & \multicolumn{1}{c}{\textgreater{}1} & \multicolumn{1}{c}{\textgreater{}3} & \multicolumn{1}{c}{\textgreater{}5} & \multicolumn{1}{c}{MAE} & \multicolumn{1}{c|}{RMS} & \multicolumn{1}{c}{\textgreater{}1} & \multicolumn{1}{c}{\textgreater{}3} & \multicolumn{1}{c}{\textgreater{}5} & \multicolumn{1}{c}{MAE} & \multicolumn{1}{c|}{RMS} & \multicolumn{1}{c}{\textgreater{}1} & \multicolumn{1}{c}{\textgreater{}3} & \multicolumn{1}{c}{\textgreater{}5} & \multicolumn{1}{c}{MAE} & \multicolumn{1}{c}{RMS} \\ \toprule \hline

\multirow{4}{*}{\rotatebox[origin=c]{90}{RGB}}&PSMNet & 65.09 & 30.87 & 13.13 & 2.54 & 4.03 & 63.62 & 28.51 & 10.40 & 2.45 & 4.26 & 63.83 & 28.41 & 10.00 & 2.43 & 4.11 \\
& PSMNet-ft & 92.67 & 31.45 & 21.32 & 4.33 & 7.76 & 92.92 & 31.24 & 20.14 & 4.13 & 7.32 & 93.24 & 30.64 & 19.65 & 4.11 & 7.43 \\
& DispNet-CSS & \tb{24.80} & 8.54 & 5.59 & 1.44 & 4.02 & 25.16 & \tb{8.47} & \tb{5.50} & \tb{1.43} & \tb{3.92} & 24.79 & \tb{8.29} & \tb{5.34} & \tb{1.38} & \tb{3.76}\\
& DispNet-CSS-ft & 39.02 & 21.12 & 14.47 & 2.37 & 4.85 & 42.29 & 21.55 & 14.28 & 2.43 & 4.88 & 40.21 & 20.91 & 14.43 & 2.40 & 4.88 \\ \hline
\multirow{4}{*}{\rotatebox[origin=c]{90}{Gray}}&PSMNet & 61.57 & 28.19 & 10.15 & 2.40 & 4.18 & 65.08 & 30.64 & 12.64 & 2.52 & 4.05 & 65.39 & 30.59 & 12.74 & 2.52 & 4.00 \\
& PSMNet-ft & 93.10 & 31.08 & 20.02 & 4.10 & 7.20 & 92.90 & 31.14 & 20.76 & 4.16 & 7.36 & 92.94 & 31.01 & 20.94 & 4.20 & 7.61 \\
& DispNet-CSS & 25.09 & \tb{8.51} & \tb{5.54} & \tb{1.43} & \tb{3.98} & \tb{25.06} & 8.62 & 5.67 & 1.46 & 4.01 & \tb{24.69} & 8.55 & 5.55 & 1.41 & 3.88 \\
& DispNet-CSS-ft & 38.83 & 20.71 & 13.81 & 2.33 & 4.77 & 42.93 & 21.76 & 14.50 & 2.45 & 4.90 & 39.93 & 21.07 & 14.54 & 2.40 & 4.87 \\
\toprule
\end{tabular}
}


\caption{\tb{Quantitative comparison between using RGB and grayscale input image for stitching conventional stereo.}}
\label{tab:quan}
\vspace{-1em}
\end{table*}

\begin{figure*}[hbt!]
\centering
    \includegraphics[keepaspectratio=true,width=\textwidth]{omnithings}

\end{figure*}

\begin{figure*}[hbt!]
\centering
    \includegraphics[keepaspectratio=true,width=\textwidth]{omnihouse}
    \caption{\tb{OmniThings and OmniHouse} From left: input fisheye images, reference panorama image, and ground truth inverse depth map.}
    \label{fig:omnihouse}
\end{figure*}

\pagebreak

{\small
\bibliographystyle{ieee}
\bibliography{egbib}
}